# К вопросу конструирования рейтингового показателя путем агрегирования частных индикаторов

*В.В. Кромер*

0. Введение в НГПУ рейтинговой системы предполагает оценивание студента по единому комплексному показателю – рейтинговому баллу или сокращенно рейтингу. На основе рейтинга будут приниматься ответственные решения (переводы, назначение стипендий, отчисления), что предполагает научную обоснованность методики расчета данного показателя.

Существующие методики основываются на взвешенном суммировании отдельных частных показателей (тестовых баллов, количеств посещенных занятий, баллов за активность на занятиях). Ниже будет показано, что данная методика абсолютно произвольна, а итоговый рейтинговый балл имеет невысокую валидность.

Дело усугубляется невысоким качеством применяемых тестов (т.н. "home-made" – «учительских» тестов). Одна из причин невысокого качества тестов – требования администраторов всех уровней, от школьных завучей до председателей аттестационных комиссий, заключающееся в том, что успевающим считается учащийся, верно выполняющий 75% заданий теста. Данное требование абсолютно безграмотно с точки зрения тестовой теории. Задания в нормативно-ориентированном тесте (а только такими и возможно объективно оценить и ранжировать студентов), подбираются исходя из 50%-выполнимости их средним студентом, соответственно зачетная норма (норма «тройки») должна составлять 25%. 75% заданий теста способны выполнить лишь 7–8% всех студентов, на которых рассчитан данный тест. Тесты с иным распределением индексов трудностей заданий невалидны и их применение в учебном процессе не оправдано. Так, аттестационное тестирование зачастую проводится тестами, полностью выполняемыми несколькими (до 5–8) студентами в группе. Ясно, что такой тест не в состоянии дифференцировать студентов и тем более не отвечает требованиям рейтинговой системы.

Ниже излагается концепция выстраивания в НГПУ рейтинговой системы (вернее, ее научного обоснования) в форме как кратких тезисов, так и развернутого обоснования отдельных пунктов предлагаемой методики. Автор осознает всю непривычность предлагаемой системы подсчета рейтинга и трудности с ее внедрением и доведением ее основных принципов до исполнителей. Идейная платформа концепции – научные воззрения Б. Мандельброта, Н. Хомского и Дж. Миллера, А.И. Яблонского, С.Д. Хайтуна. Противник тестирования в его современном виде С.Д. Хайтун [Хайтун С.Д. Нужно ли Россию тестировать // Новая газета, приложение "Наука", № 8, сентябрь 1998 г.] тем не менее видит выход, заключающийся в "развертывании" применяемых шкал и ограничении используемых шкал "от-крытыми". Автору настоящей записки, с его точки зрения, удалось довести до завершения (относительного) высказанные ранее другими исследователями предположения и догадки.

Концепция излагается последовательно по степени важности отдельных подлежащих внедрению мер.

1. С целью улучшения тестологических характеристик теста (дисперсии баллов, в частности) и внешней валидности, принять за основу предложение автора по заполнению тестовой матрицы и подсчету тестовых баллов и характеристик заданий [Кромер В.В. О некоторых вопросах тестовых технологий // Тез. докл. Второй Всеросс. конфер. "Развитие системы тестирования в России", г. Москва, 23-24 ноября 2000 г. Ч. 4. М: Прометей, 2000. С. 59-61; Кромер В.В. Калибирирование критериально-ориентированных локально-стандартизованных тестов // Вопросы тестирования в образовании. 2002. № 8. (В печати)].

2. Необходимо отказаться от применения тестов, дающих резко скошенное (левостороннее) распределение тестовых баллов. До применения теста необходимо определить генеральную совокупность, для тестирования которой предназначен данный тест (например, курс или поток). Процедура определения пригодности тестового результата (после тестирования) формализуется определением коэффициента скошенности распределения тестовых баллов. Если определение скоса не предусмотрено специальной программой обработки, это делается в программе MS Excel. При отрицательном скосе, превышающем по модулю, скажем, 1, результаты тестирования в зачет не принимаются ввиду их невалидности и невысокой дифференцирующей способности на подвыборке сильных студентов. Сказанное не относится к тестированию нерепрезентативных выборок из генеральной совокупности (например, учебной группы), где скошенность определяется составом группы, либо при тестировании стандартизированными внешними (проверочными) тестами. При левосторонней скошенности тестовых результатов с целью улучшения дифференцирующей способности теста рекомендовать последующее «дотестирование» несколькими (5–6) вопросами повышенной трудности с обработкой суммарного результата.

3. Латентная переменная, подлежащая определению, непосредственно не наблюдаема. Далее эта латентная переменная будет называться "успешность обучения" и обозначаться за $L$. Оценивается она по некоторым индикаторам. В данном предложении рассматривается работа с двумя индикаторами по каждому учебному предмету: тестовому (текущему или итоговому) баллу и количеству посещенных занятий. Активность на занятиях (например, выступление с докладом на семинаре) учитывается соответствующим увеличением индикатора "количество занятий". Взвешенное суммирование данных двух индикаторов невозможно, поскольку: а) их распределения не совпадают по форме; б) ни одно из индикаторных распределений не совпадает по форме с латентным распределением. В част-ности, распределение тестовых баллов (даже в идеальном случае его симметричности) обладает коротким хвостом в отличие от длиннохвостового распределения латентной переменной. Деформация распределений индикаторных переменных в сравнении с латентными переменными делает данные неаддитивными и последующее взвешенное суммирование некорректным. Необходимо работать с соответствующим образом нелинейно преобразованными индикаторными данными (что допустимо лишь в случае, если предыдущая деформация латентной переменной не являлась необратимой), при этом форма преобразованных распределений должна совпадать с формой латентной переменной. Поскольку латентная переменная явно не наблюдаема, форма ее распределения также неизвестна. Однако возможны некоторые предположения относительно как формы распределения латентной переменной, так и возможного математического формализма преобразования индикаторных баллов.

Большинство социальных распределений, к которым относится и распределение латентной переменной «успешность обучения», длиннохвостово и ципфово, т.е. при больших значениях переменной асимптотически стремится к распределению Ципфа. Контингент студентов НГПУ не является репрезентативной выборкой всего человечества. В НГПУ не принимают олигофренов, а исключительные по умственным способностям выпускники школ в массовом порядке в педагогические вузы не стремятся. Есть основания предполагать, что на «чистой» выборке распределение латентной переменной соответствует распределению Ципфа. Согласно ранговой форме закона Ципфа значение переменной $L$ равно

$$L \sim \frac{1}{r^{1/\alpha}}, \qquad (1)$$

где $\alpha$ – показатель распределения Ципфа, а $r$ – ранг объекта измерения в ряду, ранжированном по убывающим значениям переменной. При этом, безусловно, предполагается, что все человечество изучало, в качестве примера, практическую грамматику французского языка по принятой в НГПУ методике. Поскольку в реальности знание практической грамматики мы оцениваем на обрезанной и неравномерно прореженной выборке генеральной совокупности (всего человечества), значения рангов становятся условными и отражающими степень прореженности генеральной совокупности при формировании тестируемой выборки (студенты соответствующего потока НГПУ).

Принято считать, что коэффициент $\alpha$ отражает степень «творческости» измеряемой переменной и для ципфовых выборок $0 < \alpha < 2$, при этом $\alpha$ уменьшается с ростом «творческости». Примеры: распределение числа ученых по числу статей характеризуется значением $\alpha = 1{,}5$; распределение ученых СО АН СССР по объему публикаций характеризуется $\alpha = 1{,}3$. При подсчете рейтингового показателя необходимо задаться значением $\alpha$ для распределения исследуемой латенты. Возможно оценочное определение $\alpha$ для некоторых латент, а именно тех, где результат обучения измеряем с использованием пропорциональной шкалы (наиболее мощной из всех типов шкал). Так, возможно прямое измерение словарного запаса на иностранном языке, поскольку словарный запас имеет естественное начало отсчета 0 и открыт сверху. Измеренное значение $\alpha$ по аналогии распространяется на схожие виды учебной деятельности. Погрешность, вносимая неточным знанием этого значения, менее погрешности, вносимой пропорциональным учетом тестового балла (который, при отказе от предлагаемой методики или аналогичной, деформирован относительно измеряемого свойства). Еще более деформирована 4-балльная шкала вузовских отметок. Всем понятно, что "двоечник" не знает 50% того, что знает "хорошист", однако на практике принято обращаться со школьными отметками (и некорректно конвертированными вузовскими оценками) как с аддитивными величинами. Во всех примерах данной записки полагается $\alpha = 1,0$ (наиболее типичное значение показателя распределения Ципфа).

Нам известна одна из возможностей установить степень прореженности интересующей нас выборки из генеральной совокупности. Известно, что успешность обучения иностранному языку значимо коррелирует с коэффициентом интеллекта IQ (коэффициент корреляции $r = 0,46$) [Пимслер П. Тестирование способностей к иностранным языкам // Методика преподавания иностранных языков за рубежом. Вып. II. М., 1976. С. 168]. Поскольку распределение IQ на человечестве хорошо изучено, необходимо изучить распределение IQ на интересующей нас выборке. Полученным на выборке значениям IQ необходимо поставить в соответствие квантили (например, процентили) тех же значений IQ на генеральной совокупности. Эти квантили и будут условными рангами $r_{усл}$ значений латентной переменной на той же выборке, что позволит рассчитать (в условных единицах) значения латентной переменной (аддитивной в отличие от индикаторных баллов).

Необходимо отметить следующее. Автор не утверждает, что успеваемость определяется коэффициентом интеллекта. Делается гораздо более слабое утверждение, что в силу некоторой общности латентной переменной "интеллект", представленной индикатором "коэффициент IQ" и латентной переменной "успешность обучения", представленной индикаторами "тестовый балл" и "количество посещенных занятий" (социального характера обоих видов деятельности), обоим распределениям свойственно одинаковое распределение условных рангов. При переходе же от латент к индикаторам условные ранги, в отличие от самих значений переменных, сохраняются ввиду монотонного характера деформации. Выборка, на которой производится измерение IQ, должна репрезентировать выборку, для членов которой и определяется показатель «успешность обучения». Еди-ножды определенное распределение IQ используется в течение длительного времени для большинства учебных предметов.

4. Приведем пример определения $L$ на выборке в 10 человек (пример условный и не отражает результаты никакого реального эксперимента). В реальности выборка должна быть гораздо больше. Пилотное измерение IQ на выборке, репрезентирующей исследуемую, дало следующие результаты (Таблица 1).

Таблица 1

| № | 1 | 2 | 3 | 4 | 5 | 6 | 7 | 8 | 9 | 10 |
|---|---|---|---|---|---|---|---|---|---|---|
| IQ | 105 | 110 | 114 | 117 | 119 | 122 | 124 | 127 | 131 | 133 |
| Процентиль | 37,73 | 26,60 | 19,08 | 14.40 | 11,75 | 8,46 | 6,68 | 4,58 | 2,63 | 1,96 |
| $r_{усл}$ | 3773 | 2660 | 1908 | 1440 | 1175 | 846 | 668 | 458 | 263 | 196 |

Условные ранги $r_{усл}$ получены умножением процентиля на 100 (исходя из удобства работы с целыми числами и для сохранения точности).

Распределение IQ принято нормальным со средним значением 100 и стандартным отклонением 16. Возможно определение условных рангов и по табулированным значениям плотности распределения IQ.

Итоговое тестирование выявило следующие первичные тестовые баллы $T_б$ (Таблица 2). Для простоты количество студентов в примере выбрано равным количеству студентов при пилотном измерении IQ (Таблица 1). В общем случае при несовпадении размеров 2-х выборок производится пересчет рангов путем линейной интерполяции.

Таблица 2

| Студент | А | Б | В | Г | Д | Е | Ж | З | И | К |
|---|---|---|---|---|---|---|---|---|---|---|
| $T_б$ | 6 | 12 | 15 | 17 | 19 | 20 | 22 | 25 | 27 | 33 |
| $r_{усл}$ | 3773 | 2660 | 1908 | 1440 | 1175 | 846 | 668 | 458 | 263 | 196 |
| $L'_1$ | 27 | 38 | 52 | 69 | 85 | 118 | 150 | 218 | 380 | 511 |
| $L_1$ | 21 | 43 | 62 | 80 | 102 | 115 | 148 | 213 | 273 | 571 |

Значения $L'_1$ вычисляем по формуле (1) при $\alpha = 1,0$ путем деления произвольной величины 100000 на $r_{усл}$.

Построим график зависимости переменной $L'_1$ от тестового балла $T_б$ (рис. 1, непрерывная линия).

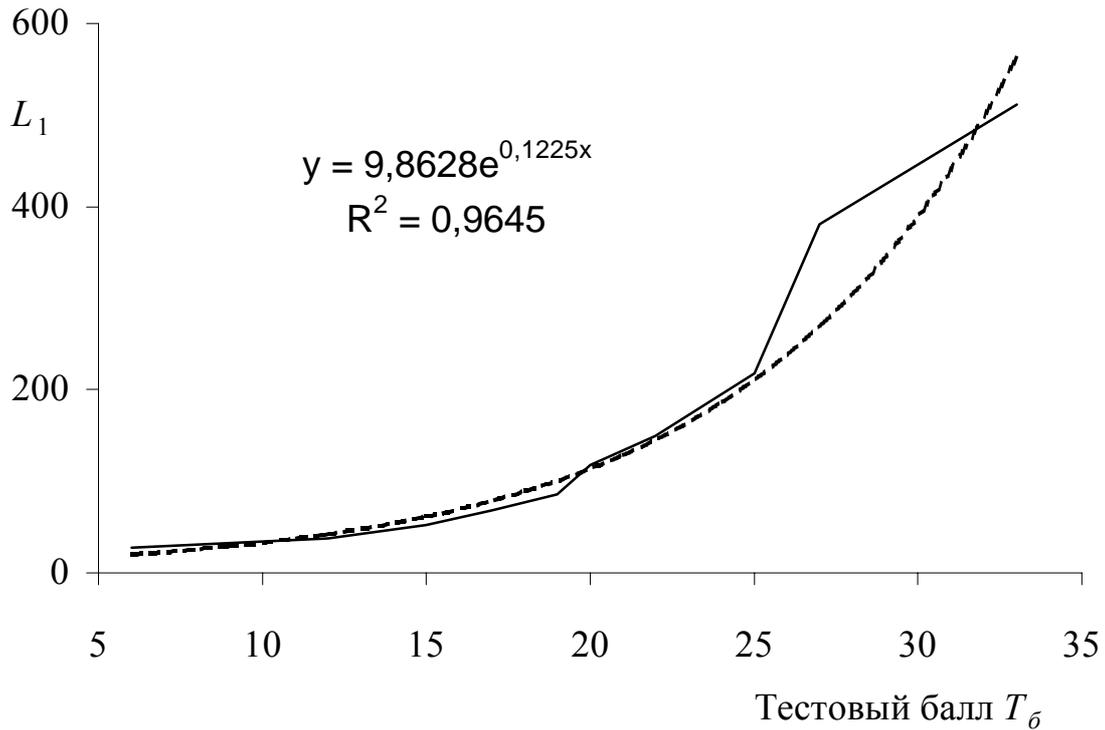

Рис. 1.

Эмпирическая зависимость $L_1$ от $T_б$ аппроксимирована теоретической зависимостью

$$L_1 = 9{,}86\, e^{0{,}123\, T_б} \qquad (2)$$

(штриховая линия). В соответствии с этой формулой производится расчет аддитивной переменной $L_1$ – одной из двух составляющих конструкта – операционального референта латентной переменной «успешность обучения» $L$. Данные занесены в таблицу 2.

Следует обратить внимание на удобство аппроксимации зависимости $L_1(T_б)$ экспоненциальной функцией. При линейном преобразовании тестового балла $T_б$ (что является вполне корректной операцией), $T_б' = c_1 T_б + c_2$, вид зависимости (2) не меняется при подстановке $T_б'$ вместо $T_б$. (Меняются лишь значения коэффициентов в формуле (2)). При выборе других аппроксимаций (например, степенными полиномами) данная устойчивость вида функции относительно линейного преобразования аргумента не сохраняется.

Рассматриваем данные по посещаемости занятий (Таблица 3). Количество посещенных занятий обозначено за $N$.

Таблица 3

| Студент | А | Б | В | Г | Д | Е | Ж | З | И | К |
|---|---|---|---|---|---|---|---|---|---|---|
| $N$ | 3 | 4 | 6 | 4 | 3 | 9 | 5 | 7 | 6 | 6 |

Ранжируем студентов по посещаемости (Таблица 4) и повторяем процесс. При совпадении $N$ для ряда студентов порядок их следования безразличен.

Таблица 4

| $N$ | 3 | 3 | 4 | 4 | 5 | 6 | 6 | 6 | 7 | 9 |
|---|---|---|---|---|---|---|---|---|---|---|
| Студент | А | Д | Б | Г | Ж | В | И | К | З | Е |
| $r_{усл}$ | 3773 | 2660 | 1908 | 1440 | 1175 | 846 | 668 | 458 | 263 | 196 |
| $L'_2$ | 27 | 38 | 52 | 69 | 85 | 118 | 150 | 218 | 380 | 511 |
| $L_2$ | 35 | 35 | 57 | 57 | 94 | 154 | 154 | 154 | 252 | 676 |

Построим график зависимости переменной $L'_2$ от количества посещений $N$ (рис. 2):

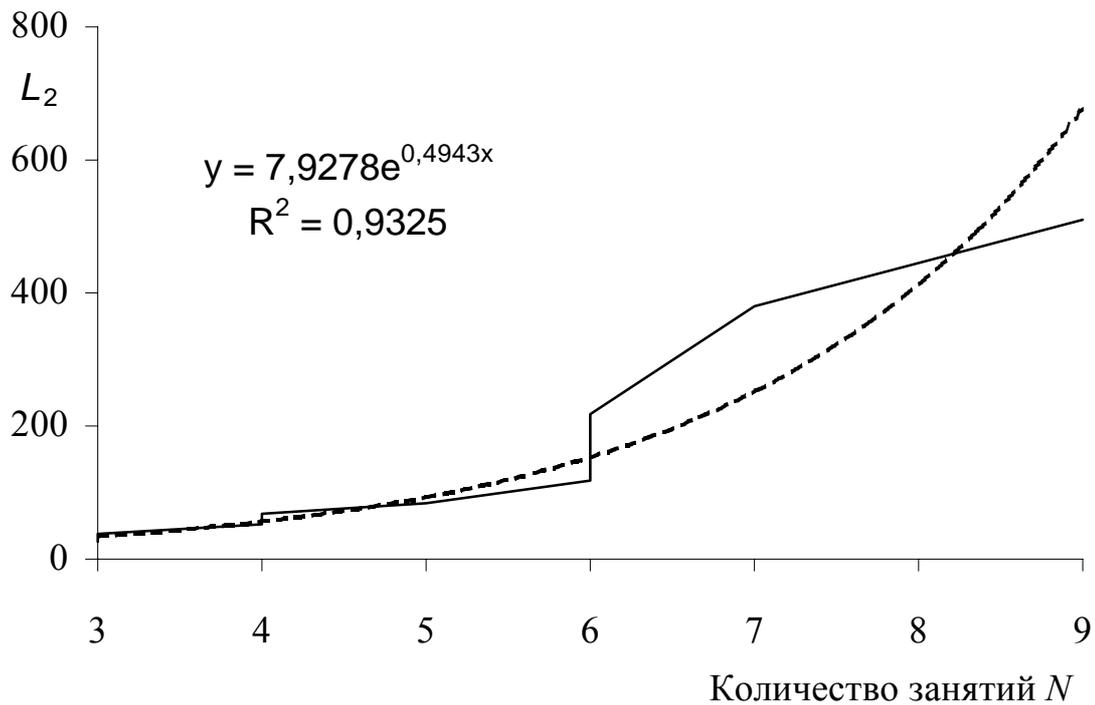

Рис. 2.

Эмпирическая зависимость $L'_2$ от $N$ (сплошная линия) аппроксимирована теоретической зависимостью

$$L_2 = 7{,}93\, e^{0{,}494\, N} \qquad (3)$$

(штриховая линия). В соответствии с этой формулой производится расчет аддитивной переменной $L_2$ – второй составляющей конструкта – операционального референта латентной переменной «успешность обучения» $L$. Данные занесены в таблицу 3.

Суммарный рейтинг $R$ студента определится взвешенной суммой двух индикаторов (теперь эта операция вполне корректна, поскольку индикаторы, как и соответствующие им латенты, аддитивны).

$$R = (1-k)L_1 + kL_2. \qquad (4)$$

Коэффициент $k$ отражает вес индикатора «посещаемость» в конструкте $R$ – операциональном референте латентной переменной «успешность обучения». При введении рейтинговой системы $k$ должно быть значительным (до 0,25-0,35), что будет стимулировать посещаемость занятий и активность на занятиях. По мере приближения обучения в НГПУ к академическим стандартам $k$ должно уменьшаться вплоть до 0. Значение $k$ устанавливается комиссией экспертов на каждый учебный год, возможно варьирование его значения по факультетам.

Данные, рассчитанные исходя из $k = 0{,}33$, занесены в таблицу 5. Именно этими данными и необходимо руководствоваться при ранжировании студентов с целью принятия управленческих решений на основе успешности обучения.

Таблица 5

| Студент | А | Б | В | Г | Д | Е | Ж | З | И | К |
|---|---|---|---|---|---|---|---|---|---|---|
| $L$ | 26 | 48 | 93 | 72 | 80 | 302 | 130 | 226 | 233 | 432 |

5. Возникает вопрос, допустимо ли обращаться с индикатором «посещаемость» аналогично индикатору «тестовый результат». Однако лишь соответствие 2 индикаторных распределений по форме обеспечивает возможность конструирования на их базе нового конструкта путем их суммирования. Зависимость успеваемости от посещаемости в общем случае неизвестна. Возможны следующие основные варианты (рис. 3), а также целый ряд не отображенных на рисунке промежуточных.

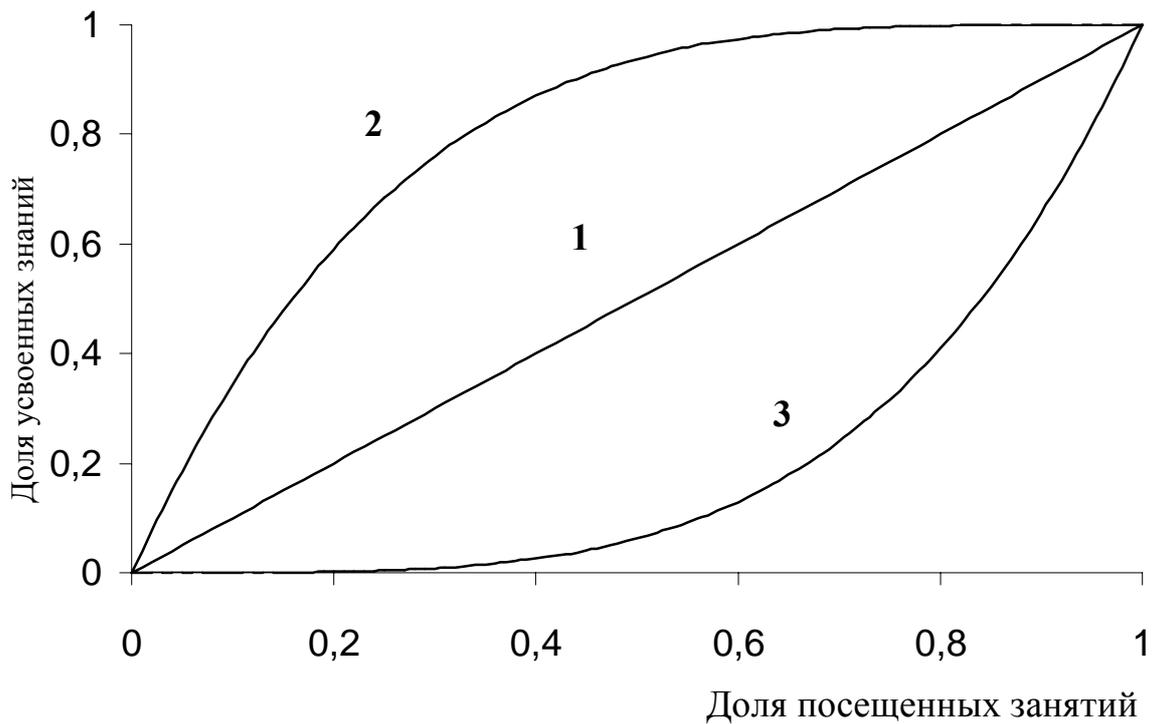

Рис. 3.

**Зависимость в соответствии с кривой 1.** Успеваемость прямо пропорциональна посещаемости. Потенциально реализуется в условиях, когда занятия являются единственным источником информации по курсу и успешность обучения напрямую зависит от количества посещенных занятий. В современных условиях (наличие учебников, распространение ксерокопирования, обмен конспектами) данная зависимость не имеет под собой основы в условиях рутинного обучения по стандартным программам, однако именно она неявно заложена в существующие «пропорциональные» рейтинговые системы. В то же время, возможно, именно подобная зависимость характерна для «штучного» обучения на очень высоком уровне. Так, значительная доля Нобелевских лауреатов являлись учениками Нобелевских же лауреатов.

**Зависимость в соответствии с кривой 2.** Реализуется в условиях, когда для успешности усвоения курса достаточно вжиться в атмосферу курса, усвоить основные требования и прочее (для чего достаточно нескольких посещений), в то время как проблем с источниками информации по фактическому наполнению курса нет.

**Зависимость в соответствии с кривой 3.** Изучаемый курс не укладывается в привычную систему понятий. Даже к середине курса студент еще не в состоянии уложить знания в систему. Лишь к концу курса наступает прояснение и резкое возрастание «успешности».

В условиях, когда истинный вид зависимости неизвестен, и эти зависимости отличаются для разных учебных предметов и варьируют от студента к студенту, приемлема наиболее общая (инвариантная) методика выравнивания форм распределений.

Полученный взвешенным суммированием (сверткой или агрегированием) конструкт «успешность обучения по предмету» также аддитивен, что позволяет использовать его в дальнейших свертках для получения операционального референта латентной переменной «успешность обучения по всем предметам» путем взвешенного суммирования конструктов по отдельным предметам. В качестве весов отдельных предметов при этом берется количество часов на соответствующие курсы по учебному плану либо определенное экспертным методом число.

6. В условиях рейтинговой системы применение стандартной четырехбалльной шкалы отметок нерационально ввиду грубости и низкой дифференцирующей силы данного индикатора. Однако в случае вынужденного использования отметок в качестве индикатора также возможно установление зависимости между латентной переменной и индикатором аналогично примерам разделов 2, 3.

7. Описанная система в отличие от используемых имеет когнитивное обоснование и предположительно ранжирует студентов более объективно. Все вычисления производятся по единожды составленной программе. Несмотря на худшую прозрачность системы по сравнению с "пропорциональными" балльными системами, конечный результат, как высоко коррелирующий с интуитивно воспринимаемыми оценками и самооценками студентов, будет положительно воспринят ими.